\documentclass[11pt]{article}

\usepackage[final]{acl}
\usepackage{multirow}
\usepackage{booktabs}
\usepackage{listings}
\usepackage{times}
\usepackage{latexsym}
\usepackage{amsmath}
\usepackage{hyperref}
\usepackage{graphicx}
\usepackage{subcaption}
\usepackage[T1]{fontenc}

\usepackage[utf8]{inputenc}

\usepackage{microtype}

\usepackage{inconsolata}
\usepackage{amssymb}

\usepackage{graphicx}

\newcommand{\postspace}{\vskip -3mm}
\newcommand{\minipostspace}{\vskip -2mm}

\title{A Dynamic Self-Evolving Extraction System}

\author{Moin Amin-Naseri \\
  Megagon Labs \\
  \texttt{moin@megagon.ai} \\\And
  Hannah Kim \\
  Megagon Labs \\
  \texttt{hannah@megagon.ai} \\\And
  Estevam Hruschka \\
  Megagon Labs \\
  \texttt{estevam@megagon.ai} \\}

\definecolor{cadmiumgreen}{rgb}{0.0, 0.42, 0.24}
\definecolor{cardinal}{rgb}{0.77, 0.12, 0.23}
\definecolor{cadmiumred}{rgb}{0.89, 0.0, 0.13}

\usepackage[most]{tcolorbox}
\newtcolorbox[list inside=prompt,auto counter,number within=section]{prompt}[1][]{
    fontupper=\ttfamily\footnotesize,
    boxsep=5pt,
    left=0pt,
    right=0pt,
    top=0pt,
    bottom=0pt,
    boxrule=1pt,
    breakable,
    #1,
}

\begin{document}
\maketitle

\begin{abstract}
The extraction of structured information from raw text is a fundamental component of many NLP applications, including document retrieval, ranking, and relevance estimation. High-quality extractions often require domain-specific accuracy, up-to-date understanding of specialized taxonomies, and the ability to incorporate emerging jargon and rare outliers. In many domains--such as medical, legal, and HR--the extraction model must also adapt to shifting terminology and benefit from explicit reasoning over structured knowledge. 
We propose DySECT, a Dynamic Self-Evolving Extraction and Curation Toolkit, which continually improves as it is used. The system incrementally populates a versatile, self-expanding knowledge base (KB) with triples extracted by the LLM. The KB further enriches itself through the integration of probabilistic knowledge and graph-based reasoning, gradually accumulating domain concepts and relationships. The enriched KB then feeds back into the LLM extractor via  prompt tuning, sampling of relevant few-shot examples, or fine-tuning using KB-derived synthetic data. As a result, the system forms a symbiotic closed-loop cycle in which extraction continuously improves knowledge, and knowledge continuously improves extraction\footnote{Code and installation provided at: \url{https://github.com/megagonlabs/dysect}}.
\end{abstract}

\section{Introduction}

Information extraction (IE) is a core component of modern NLP pipelines, supporting tasks such as document retrieval, ranking, and knowledge base population across domains where terminology and schemas evolve rapidly \citep{jehangir2023ner_survey,xu2024large,zhang2025survey}. Recent works on generative IE and LLM-based extraction highlight a shift from token-level classifiers to sequence-to-structure generation, enabling unified handling of named entities, relations, and events across diverse formats \citep{xu2024large,zhang2025survey}. At the same time, domain-specific case studies, for example in scientific and clinical text, show that high-quality extraction still requires careful schema design, substantial domain adaptation, and rigorous evaluation \citep{dagdelen2024structured,adam2025clinical_ie}. Taken together, these lines of work suggest that both classic neural IE systems and newer LLM-based approaches remain heavily dependent on curated datasets and manually engineered adaptation strategies.

Despite progress in label-efficient and continual learning methods, updating IE systems in response to new data remains cumbersome. Self-training and bootstrapping approaches reduce annotation costs but typically rely on manually designed seed rules, confidence thresholds, and offline retraining cycles \citep{angeli2015bootstrapped,sarkhel2023self}. Domain-adaptive continual learning for LLMs aims to incrementally absorb new domain data without catastrophic forgetting, yet it generally assumes explicit training phases, access to model weights, and non-trivial engineering to maintain stability \citep{rostami2024continuous,kharrat2023cl_survey}. Frameworks that combine ontologies or knowledge bases with language models can improve robustness and interpretability \citep{salvador2025review}, but they are often pipeline-based and still depend on human-engineered schemas, rather than forming a simple, closed feedback loop in which using the extractor directly and continuously refines the underlying knowledge and, in turn, improves future extractions.

In this work, we present DySECT---Dynamic Self-Evolving Extraction \& Curation Toolkit---a dynamic, self-improving extraction framework that directly addresses these limitations (Figure~\ref{fig:system-overview}). This idea is related to earlier work on continuously learning systems such as Never-Ending Language Learning (NELL) \cite{mitchell2018never}, which introduced the idea of continuously extracting and promoting knowledge into an evolving knowledge base through iterative learning over large-scale web data.

Rather than relying on discrete retraining cycles, curated adaptation data, or access to model weights, our system improves simply through use. Each extraction call produces structured triples that are inserted into a self-evolving knowledge base (KB) equipped with probabilistic confidence modeling. The KB assigns reliability scores to stored triples based on source credibility and relation frequency, enabling it to maintain a continuously updated view of domain concepts without requiring explicit supervision or ontology curation. Crucially, this knowledge is then fed back into the extractor through prompt augmentation, KB-informed concept abstractions, and optionally synthetic data generation for lightweight fine-tuning. This closed-loop design allows the extractor to become increasingly domain-aware, consistent, and accurate over time, transforming extraction from a static prediction task into an iterative knowledge acquisition process.

\section{DySECT}

\begin{figure}[t]
  \centering
  \includegraphics[width=\linewidth]{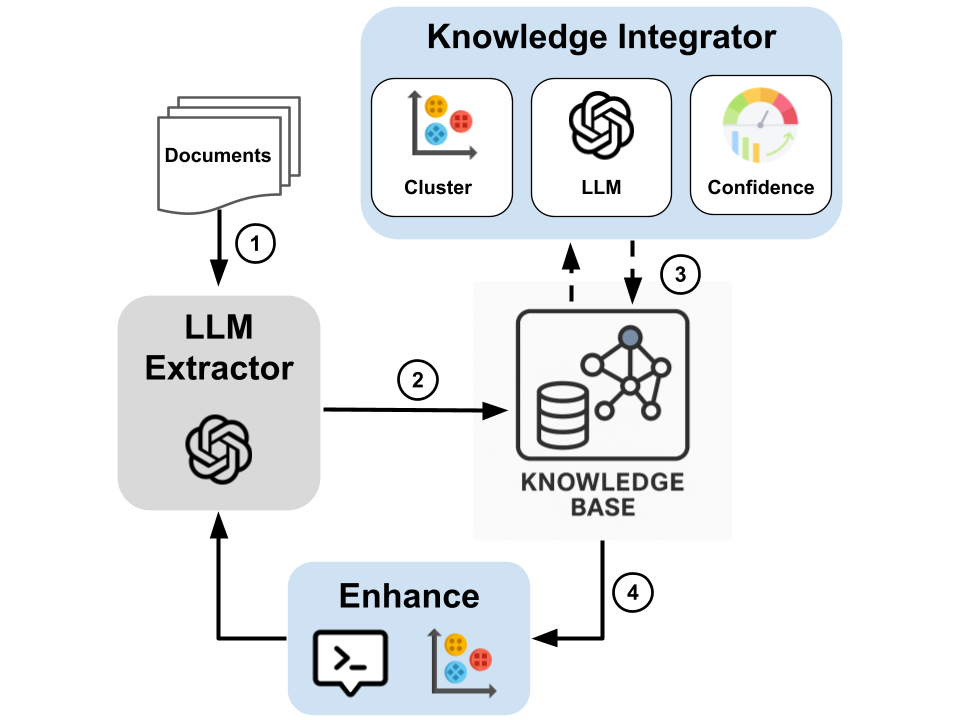}
  \caption{Overview of our closed-loop extraction and knowledge base system. 
  The extractor produces triples from raw text, which populate the KB and 
  are then fed back to improve future extraction.}
  \label{fig:system-overview}
\end{figure}

Our system consists of three major components (see Figure~\ref{fig:system-overview}):

\begin{enumerate}
    \item \textbf{Extraction Step:} An LLM prompted to produce concept-level triples from raw text.
    \item \textbf{Knowledge Base Growth:} An evolving graph that stores triples, integrates newly acquired knowledge, and supports hierarchical reasoning as well as mutual-exclusivity reasoning, using both to estimate confidence scores for candidate facts.
    \item \textbf{Feedback Mechanisms:} Methods to re-inject KB knowledge into the extractor via prompts, examples, or fine-tuning.
\end{enumerate}

The core idea is a closed-loop design: extraction populates the KB and the KB improves future extraction.


\subsection{Extraction Step}

Given a modifiable prompt~\ref{prompt:task-description}, the LLM extracts triples of the form:

\[
(\text{subject}, \text{relation}, \text{object})
\]

After adjusting these triples to the proper format, they are inserted directly into the KB.

\subsection{Knowledge Base Growth}
\label{sec:knowledgeBaseGrowth}
The KB growth happens based on two nested loops. For each extraction task, the KB is initially empty and the \textbf{outer loop} iterates over batches of the extractor outputs. In each outer iteration, the system ingests triples produced by the extractor, assigning them a source label (e.g., \texttt{Extractor\_gpt-4o}). 

Within each extractor batch, the system performs a few iterations (\textbf{inner loop}) of further content acquisition. Each inner iteration consists of:

\paragraph{Knowledge integration (KB update).}
    The system calls a \texttt{knowledgeIntegrator} module to consolidate newly added evidence and enforce lightweight ontology constraints, including mutual-exclusivity constraints and overall confidence. 
    The knowledge integrator also structures the KB's  concepts into more interpretable hierarchies. For any node with a large or semantically heterogeneous set of children, we apply KNN-based clustering over the children’s embeddings to group conceptually similar nodes into coherent clusters. A small representative sample from each cluster is then provided to an LLM, which proposes a concise label summarizing the shared semantics of that group (and a short description). The generated label becomes a new intermediate node inserted between the original parent and its children, resulting in a more navigable and semantically meaningful hierarchy. These automatically discovered abstractions distill dense concept spaces into higher-level categories that can be surfaced to the extractor, improving its ability to generalize, recognize related concepts, and avoid redundant or inconsistent extractions. This step updates the KB state that will condition the next inner iteration.

\paragraph{Concept acquisition (instance discovery)}
    For each concept in the KB, the concept acquisition module retrieves KB examples\footnote{The more concept instances the KB has, the more it can further explore the acquisition of extra instances.} and prompts one or more LLMs to propose additional instances. The newly proposed instances are then written back as candidate triples with an initial confidence.

\paragraph{Relation acquisition (fact discovery).}
    For each typed relation (subjectType--predicate--objectType) in the KB, the relation acquisition module samples candidate argument pairs from the KB and prompts the same set of LLMs, used during concept acquisition, to propose new relation instances. Inverse relations are automatically created. Newly proposed relation triples are added to the KB.

These two nested loops create a continual, closed-loop cycle in which (i) extractor/LLM acquisition expands the pool of candidate facts and (ii) knowledge integration consolidates and constrains these facts, producing an improved KB state that conditions subsequent extraction.

The knowledge base representation is built on top of Theo \cite{mitchell1989theo} with an accompanying library that enables continuous expansion, organization, and reliability estimation of stored information. Each triple added to the KB is enriched with a probabilistic confidence score that reflects both the credibility of its source and the strength of cumulative evidence observed across extraction calls. 

For each triple $t=(s,p,o)$, the knowledge base stores one or more \emph{local} confidence values $c_i \in [0,1]$ produced by upstream extractors and acquisition LLMs, along with a frequency $f_i \in \mathbb{N}$ indicating repeated observations of the same content.
If the triple is associated with a source in a predefined set of fully trusted sources\footnote{In the current version, the only fully trusted sources are the human curators, who can provide feedback using an interactive interface for the knowledge base.}, we set its overall confidence to $C(t)=1.0$.
Otherwise, we aggregate local evidence using a conservative noisy-or with a shrinkage factor $\lambda \in (0,1]$ (default $\lambda=0.75$). Frequencies are incorporated by treating $f_i$ as repeated independent support for the same confidence value (equivalently, by exponentiating the corresponding term):
\begin{equation}
C_{\mathrm{agg}}(t) \;=\; 1 - \prod_i \left(1-\lambda c_i\right)^{f_i}.
\end{equation}
This aggregation increases with higher local confidences and repeated supporting evidence, while the factor $\lambda$ prevents overconfidence when evidence is noisy or redundant.

Finally, we penalize conflicts with mutual-exclusivity constraints. Let $m(t)$ be the number of mutually exclusive instances detected for $t$. The final confidence stored in the knowledge base is:
\begin{equation}
C(t) \;=\; \frac{C_{\mathrm{agg}}(t)}{m(t)+1}.
\end{equation}
This preserves accumulated support while explicitly down-weighting triples that compete with mutually exclusive alternatives. This confidence modeling allows the KB to prioritize high-quality knowledge and suppress noisy triples when generating prompt augmentations, selecting examples, or constructing synthetic training data.

In spite of DySECT's ability to autonomously evolve with no human supervision, it also provides an interactive interface for the knowledge base that provides direct transparency and control over accumulated knowledge, whenever necessary. The interface allows users to inspect stored triples, monitor confidence statistics and hierarchical structure, and explicitly validate, invalidate, or manually insert new triples when necessary. This human-in-the-loop capability enables policy enforcement, bias correction, and targeted refinement of the KB, but it does not impose mandatory human curation. Unlike fully parameterized language models, where knowledge is implicitly encoded in weights and difficult to inspect or modify directly, our framework maintains knowledge in an explicit and editable form. This design supports autonomous and iterative self-evolving as well as controlled oversight, ensuring that adaptation does not come at the cost of interpretability.

\begin{figure*}[t]
  \centering
  \includegraphics[width=\textwidth]{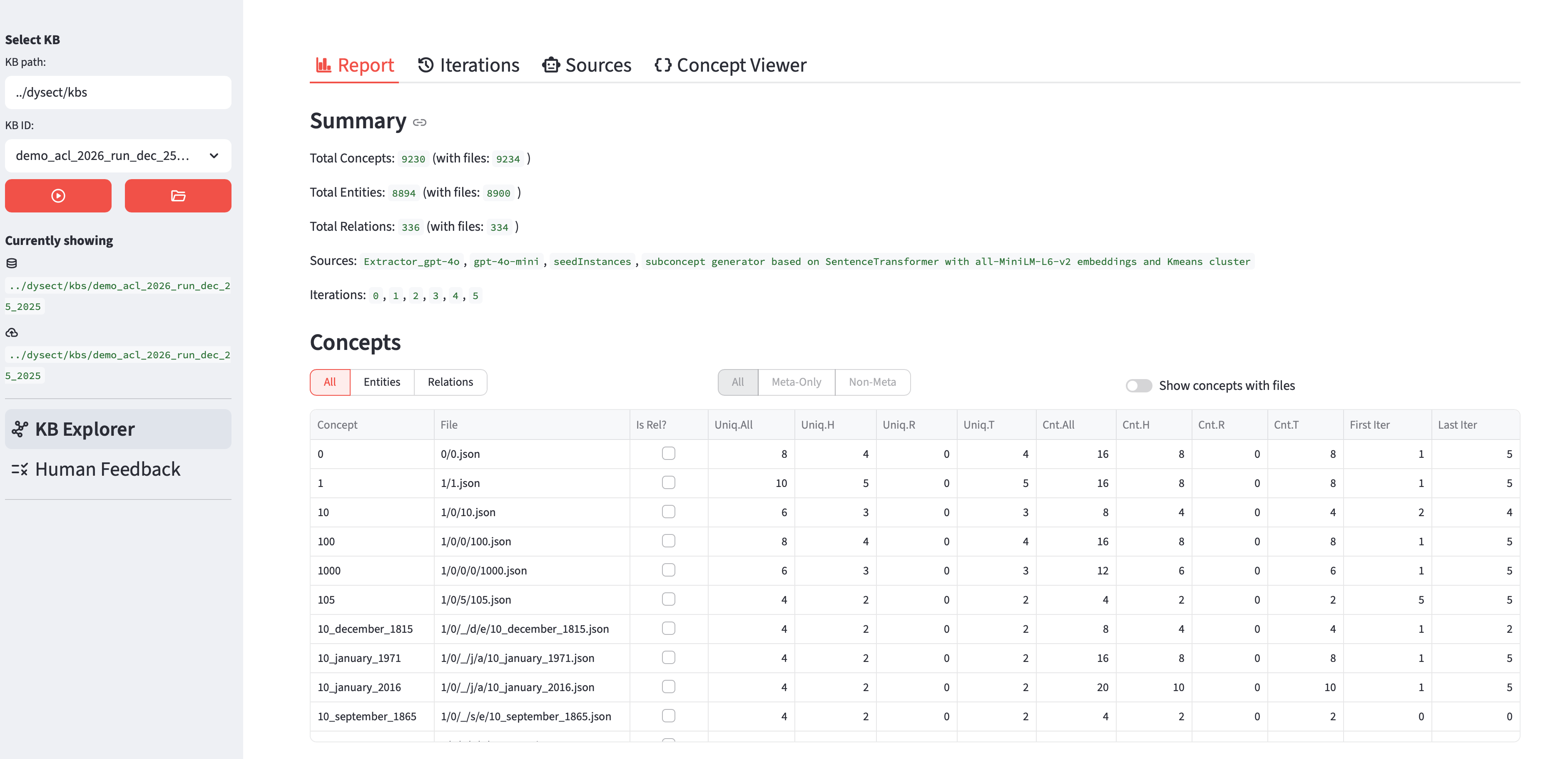}
  \caption{Knowledge base dashboard displaying summary statistics and concept analytics.}
  \label{fig:main-demo}
\end{figure*}

\subsection{Feedback Mechanisms}
\label{sec:feedbackMechanism}
To close the loop between extraction and knowledge accumulation, the KB provides several mechanisms that feed its learned structure back into the extractor. One pathway operates through prompt augmentation: the system maintains a configurable prompt template~\ref{prompt:task-description} that can be periodically updated by querying the KB for relevant high-confidence information. Depending on the extraction domain, this may include representative instances of a concept, long-tail examples with high reliability scores, or contextual groupings such as skills commonly associated with a particular job family or entity type. These retrieved snippets are injected directly into the extraction prompt~\ref{prompt:task-description}, enabling the LLM to condition on domain-aware and up-to-date knowledge without requiring explicit retraining. A second pathway leverages the hierarchical abstractions (subcategories) constructed by the KB's knowledge integrator (see section \ref{sec:knowledgeBaseGrowth}). 
These automatically discovered subcategories as well as mutually exclusive concepts and relations (also automatically created by the knowledge integrator) can be surfaced to the extractor as conceptual anchors, negative examples or category-level cues, helping it identify, group, and generalize concepts more effectively. 

Finally, 
the KB can be used to generate synthetic training data by translating its distilled structured knowledge into raw text. For triples or clusters that meet a specified confidence threshold, the system produces factual natural-language descriptions—similar to the knowledge-to-text generation techniques shown to be effective in prior work such as \citet{mruthyunjaya2023rethinking}. This synthetic corpus can be used to fine-tune the extractor or calibrate its outputs, enabling it to absorb domain knowledge encoded in the KB without requiring manually labeled datasets. Through this combination of prompt enrichment, hierarchical abstraction, and knowledge-driven text generation, the KB continuously guides and reinforces the extractor, forming a self-improving feedback loop.

\section{System Demonstration}

We demonstrate the system through a concrete extraction scenario on a Wikipedia-style document. The interface exposes how structured knowledge accumulates, organizes itself, and becomes queryable over iterations (Figure \ref{fig:main-demo}).

\paragraph{Step 1: Initial Extraction.}
We begin with a base prompt and perform a first round of extraction. The model identifies a limited set of triples, which in this case focused more on relations about "date" and "person" concepts:

\begin{lstlisting}
[
  ["Moneytalks", "producer", "Bruce Fairbairn"],
  ["Moneytalks", "publication date", "21 September 1990"],
  ["Live", "publication date", "1992"]
]
\end{lstlisting}

At this stage, the extraction captures surface-level factual relations but misses broader relational coverage such as performer associations and related musical entities. These triples are inserted into the knowledge base (KB).

\paragraph{Step 2: Knowledge Base Integration.}
We trigger the knowledge integrator with the newly inserted triples. Over several internal iterations (Figure \ref{fig:iterations}), the KB: \\
- consolidates triples and updates confidence scores, \\
- discovers hierarchical abstractions using KNN clustering, \\
- assigns cluster labels using GPT-4o-mini (Figure \ref{fig:emerg}), \\
- and identifies mutually exclusive concept families, e.g., \textit{Sports Organizations} mutually exclusive with \textit{Religious Organizations} (Figure \ref{fig:mutex}).

After these iterations, the KB contains structured, higher-level abstractions such as \textit{Music Genre: Rock}, \textit{Political Organizations}, and \textit{Awards}, forming a richer conceptual representation as opposed to general concepts like \textit{Organizations} or \textit{Miscellaneous}.

\paragraph{Step 3: KB-Guided Extraction.}
For the next extraction round, we query the KB using the concepts identified in the previous iteration (Figure \ref{fig:emerg}). The KB returns higher-level abstractions, which are injected into the prompt as guidance:

\begin{lstlisting}
### Previously Extracted General Concepts:
Music Genre: Rock, Time, Persons
\end{lstlisting}

This guidance shifts the model’s focus toward music-related relations and associated entities.

The resulting extractions show substantially improved coverage:

\begin{lstlisting}
[
  ["Moneytalks", "producer", "Bruce Fairbairn"],
  ["The Razors Edge", "performer", "AC / DC"],
  ["Moneytalks", "performer", "AC / DC"],
  ["Moneytalks", "publication date", "21 September 1990"],
  ["Live", "performer", "AC / DC"],
  ["Moneyball", "publication date", "2011"],
  ["Live", "publication date", "1992"]
]
\end{lstlisting}

Compared to the initial round, additional true positives emerge, particularly performer relations and conceptually related entities that were previously missed.

\paragraph{Observation.}
This scenario illustrates how accumulated structured knowledge influences subsequent extraction rounds. The KB does not modify model parameters; instead, it provides conceptual abstractions that steer the extractor toward richer relational coverage. As more documents are processed, the KB becomes more structured and informative, enabling continued improvement while remaining transparent and editable.

This iterative interaction between extraction and structured knowledge accumulation demonstrates how the system evolves through usage, improving coverage without explicit retraining.


\section{Experimental Setup}
\paragraph{Dataset:} We evaluate our approach on \textbf{DocRED} \citep{yao2019docred}, a large-scale document-level relation extraction benchmark constructed from Wikipedia articles. DocRED contains 3,053 training, 1,000 development, and 1,000 test documents, covering 96 relation types and a diverse set of entity categories such as person, location, organization, time, and miscellaneous entities. Since DocRED provides both supervised and weakly supervised annotations, we use only the supervised annotations throughout our experiments for higher annotation reliability.

For our iterative extraction setup, we randomly sample 500 documents from the supervised training split to serve as a fixed development set for evaluating extraction quality across successive extraction iterations. In addition, we separately sample 100 documents whose annotations are used exclusively to initialize the raw KB prior to the start of the iterative refinement process. To align with our extraction pipeline and evaluation procedure, we preprocess the DocRED annotations into TSV files containing $(subject, predicate, object)$ triples extracted from the supervised annotations. These triples are then used as the reference ground-truth set for computing extraction performance metrics throughout the iterative process.

\paragraph{Self-Evolving Loop:} Our experiments simulate the proposed self-evolving loop in three stages: (1) We first run a base LLM extractor on DocRED documents using a fixed prompt to produce triples. These triples are inserted into the knowledge base (KB) along with their associated confidence scores.

(2) Following initial triple ingestion, the KB enters an integration stage in which newly added triples are consolidated and confidence values are updated. The knowledge integrator merges semantically equivalent facts, maintains frequency statistics, and incrementally refines the graph structure. When nodes accumulate many heterogeneous children, clustering-based abstraction introduces intermediate subconcepts, forming a more organized hierarchy that can later guide subsequent extraction rounds.

(3) The enriched KB is then queried to retrieve high-confidence hierarchical subconcepts, which are fed back into the extractor prompt. This guidance is provided in two modes:

\paragraph{Feedback Modes:} We consider two feedback modes: (1) \textit{Encouraging mode (positive)}, where high-confidence subconcepts and representative relations are injected as positive examples into the prompt~\ref{prompt:Positive}, encouraging the model to extract additional instances belonging to similar concept families. (2) \textit{Prohibitive mode (negative)}, where well-covered or saturated subconcepts are explicitly marked in the prompt~\ref{prompt:Negative} as already extracted, prompting the model to avoid redundant patterns and instead focus on discovering new relations or underrepresented concept clusters.
We iterate this procedure, using prior extractions to query the increasingly enriched KB and measure further improvements.

\paragraph{Models:} We evaluate four large language models: GPT-4.1, GPT-4.1-mini, LLaMA-3.3 70B, and Kimi K2.5. Each model is evaluated under two conditions: (i) baseline extraction without KB feedback and (ii) KB-guided extraction using hierarchical subconcepts.

\section{Experimental Result}
KB-guided extraction consistently improves recall across all four evaluated models. Notably, even without synthetic data generation or fine-tuning, simply exposing the extractor to KB-derived hierarchical abstractions yields a recall improvement of 5–8\% on 1st iteration of extractions compared to the baseline configuration without KB feedback. This supports our central claim that the system improves its extraction capability purely through structured knowledge accumulation and reuse of the system.

The largest gains are observed for GPT-4.1, which benefits most from structured hierarchical guidance. This suggests that stronger reasoning models are better able to leverage abstracted subconcepts to expand coverage of relevant relations. GPT-4.1-mini, LLaMA-3.3 70B, and Kimi K2.5 also show consistent recall improvements within the same range, indicating that the effect is model-agnostic.

\begin{table}[t!]
\small
\centering
\begin{tabular}{lllrr}
\toprule
&&\bf Model&\bf Recall&\bf Avg \# Ext-Trip\\
\midrule
\multirow{4}{*}{\rotatebox[origin=c]{90}{\bf Base}}&\multirow{4}{*}{\rotatebox[origin=c]{90}{Iter-0}}&GPT 4.1 &22.80&15.34\\ 
&&GPT 4.1 mini&12.72&\bf 7.71\\
&&Kimi K2.5&\bf 32.03&20.37\\
&&LLaMA-3.3 70B &23.23&13.30\\
\midrule
\multirow{8}{*}{\rotatebox[origin=c]{90}{\bf DySECT - Positive}}&\multirow{4}{*}{\rotatebox[origin=c]{90}{Iter-1}}&GPT 4.1 &30.62&17.33\\
&&GPT 4.1 mini&20.22&\bf 13.59\\
&&Kimi K2.5&\bf 37.85&29.15\\
&&LLaMA-3.3 70B&28.78&19.77\\\cline{2-5}
\\[-1.8ex]
&\multirow{4}{*}{\rotatebox[origin=c]{90}{Iter-2}}&GPT 4.1 &37.03&27.67\\
&&GPT 4.1 mini&27.67&23.00\\
&&Kimi K2.5&\bf 44.41&34.28\\
&&LLaMA-3.3 70B&30.97&\bf 22.53\\
\midrule
\multirow{8}{*}{\rotatebox[origin=c]{90}{\bf DySECT - Negative}}&\multirow{4}{*}{\rotatebox[origin=c]{90}{Iter-1}}&GPT 4.1 &28.72&16.03\\
&&GPT 4.1 mini&17.18&\bf 12.52\\
&&Kimi K2.5&\bf 37.30&28.15\\
&&LLaMA-3.3 70B&27.79&19.18\\\cline{2-5}
\\[-1.8ex]
&\multirow{4}{*}{\rotatebox[origin=c]{90}{Iter-2}}&GPT 4.1 &31.49&26.64\\
&&GPT 4.1 mini&29.18&22.17\\
&&Kimi K2.5&\bf 43.63&32.99\\
&&LLaMA-3.3 70B&29.76&\bf 21.67\\
\bottomrule
\end{tabular}
\caption{DySECT performance on the DocRED benchmark, reported in terms of Recall and the average number of extracted triples.}
\label{tab:main_result}
\end{table}

\paragraph{Analysis of Feedback Modes.}
The two feedback modes encourage different exploration behaviors during extraction. In the positive feedback mode, the KB provides representative concepts, relations, and high-confidence examples that guide the extractor toward semantically consistent regions of the knowledge space. This reinforced guidance improves schema alignment and extraction stability over iterations, which likely explains its stronger overall performance in our experiments.

In contrast, the negative feedback mode discourages previously observed concepts or relations to promote exploration of less-covered regions of the KB. While this can introduce noisier or lower-confidence extractions, DySECT’s knowledge integration mechanism mitigates uncontrolled KB growth through confidence estimation, validation heuristics, and optional human verification. As a result, the system can continue exploring new concepts and domains while maintaining KB quality over time. This mode can therefore be particularly useful in settings where discovery and coverage expansion are prioritized, such as early-stage KB construction or rapidly evolving domains.

A promising future direction is combining both modes dynamically by balancing consolidation and exploration. For example, the system could reinforce high-confidence concepts while selectively discouraging overrepresented patterns, potentially improving both extraction stability and knowledge diversity.

These findings support our central claim: through repeated use of the proposed feedback mechanism, both the extractor and the knowledge base progressively improve without requiring explicit retraining. As extraction results accumulate, the KB becomes more structured and reliable, and the refined hierarchical abstractions further guide subsequent extraction rounds. This mutual refinement enables continuous adaptation while preserving transparency and control. Through our tool, users can inspect the accumulated knowledge, verify high-confidence triples, and intervene to correct or refine the KB when necessary. The framework therefore combines self-improvement with interpretability, allowing the system to evolve over time without sacrificing oversight.

\section{Conclusion}

We introduced a dynamic extraction framework in which structured knowledge accumulation and extraction operate in a mutually reinforcing loop. Each round of extraction enriches the knowledge base, and the resulting hierarchical abstractions are fed back to guide subsequent predictions. This process enables the system to adapt and expand coverage over time without explicit retraining.

Beyond performance gains, the framework maintains transparency and controllability: accumulated knowledge can be inspected, verified, and revised as needed. Such oversight is particularly important in domains where policy, compliance, or domain-specific constraints must be respected. Empirically, we observe consistent recall improvements across models through iterative KB-guided prompting, demonstrating that structured knowledge reuse alone can meaningfully enhance extraction performance.

\section{Limitations}

While DySECT demonstrates improvements in iterative extraction quality and KB-guided refinement, several limitations remain. First, the framework still depends on the quality of the underlying extractor and prompting strategy, meaning extraction biases or hallucinations may propagate into the KB over time. Although confidence estimation, validation heuristics, and optional human verification help mitigate this issue, they cannot fully eliminate noisy knowledge accumulation. In addition, the effectiveness of KB querying and knowledge reuse mechanisms may vary depending on the domain, data characteristics, and task setting.

Finally, iterative extraction and KB integration introduce additional computational overhead compared to one-pass extraction pipelines due to the need for repeated KB querying and integration during extraction. As the KB grows in size and complexity, maintaining efficient knowledge integration and KB integrity might add to the challenge.




\section*{Broader Impact Statement}

This work advances the development of adaptive and transparent information extraction systems that improve through structured knowledge accumulation rather than opaque parameter updates. By maintaining knowledge in an explicit, inspectable, and editable form, the framework supports traceability, auditability, and human collaboration. Such capabilities are particularly valuable in research, analytics, and enterprise knowledge management settings where understanding and controlling model behavior is essential.

The proposed approach also encourages responsible AI practices by combining autonomous improvement with built-in oversight mechanisms. The ability to monitor, validate, and refine accumulated knowledge enables safer deployment in evolving environments and supports long-term maintainability. By bridging automation with transparency, this framework offers a practical step toward more controllable and interpretable AI systems.

\section*{Ethics Statement}

Our experiments use DocRED, a publicly available dataset derived from Wikipedia and Wikidata. Because the framework accumulates and reuses its own extractions, erroneous or biased triples could be reinforced over time. Confidence modeling mitigates this risk, but human oversight remains important to verify and correct the evolving knowledge base. Responsible deployment requires domain-specific validation and monitoring.

\section*{Acknowledgements}

We would like to thank everyone who contributed to and supported this work. In particular, we are grateful to the production team, especially Shingo Kato and James Levine, for their continuous support throughout the project. We also sincerely thank the annotation team for their valuable efforts in providing high-quality ground-truth data, with special thanks to Rone Yamasaki for his support and contributions. Their collective efforts played an important role in shaping this work.

\bibliography{custom}

@inproceedings{yao2019docred,
    title = "{D}oc{RED}: A Large-Scale Document-Level Relation Extraction Dataset",
    author = "Yao, Yuan  and
      Ye, Deming  and
      Li, Peng  and
      Han, Xu  and
      Lin, Yankai  and
      Liu, Zhenghao  and
      Liu, Zhiyuan  and
      Huang, Lixin  and
      Zhou, Jie  and
      Sun, Maosong",
    booktitle = "Proceedings of the 57th Annual Meeting of the Association for Computational Linguistics",
    month = jul,
    year = "2019",
    address = "Florence, Italy",
    publisher = "Association for Computational Linguistics",
    url = "https://aclanthology.org/P19-1074",
    doi = "10.18653/v1/P19-1074",
    pages = "764--777",
}

@article{jehangir2023ner_survey,
  title={A survey on named entity recognition—datasets, tools, and methodologies},
  author={Jehangir, Basra and Radhakrishnan, Saravanan and Agarwal, Rahul},
  journal={Natural Language Processing Journal},
  volume={3},
  pages={100017},
  year={2023},
  publisher={Elsevier}
}

@article{xu2024large,
  title={Large language models for generative information extraction: A survey},
  author={Xu, Derong and Chen, Wei and Peng, Wenjun and Zhang, Chao and Xu, Tong and Zhao, Xiangyu and Wu, Xian and Zheng, Yefeng and Wang, Yang and Chen, Enhong},
  journal={Frontiers of Computer Science},
  volume={18},
  number={6},
  pages={186357},
  year={2024},
  publisher={Springer}
}

@inproceedings{zhang2025survey,
  title={A survey of generative information extraction},
  author={Zhang, Zikang and You, Wangjie and Wu, Tianci and Wang, Xinrui and Li, Juntao and Zhang, Min},
  booktitle={Proceedings of the 31st International Conference on Computational Linguistics},
  pages={4840--4870},
  year={2025}
}

@article{dagdelen2024structured,
  title={Structured information extraction from scientific text with large language models},
  author={Dagdelen, John and Dunn, Alexander and Lee, Sanghoon and Walker, Nicholas and Rosen, Andrew S and Ceder, Gerbrand and Persson, Kristin A and Jain, Anubhav},
  journal={Nature communications},
  volume={15},
  number={1},
  pages={1418},
  year={2024},
  publisher={Nature Publishing Group UK London}
}

@article{adam2025clinical_ie,
  author       = {Adam, Hammaad},
  title        = {Clinical Information Extraction with Large Language Models: A Case Study on Organ Procurement},
  journal      = {AMIA Annual Symposium Proceedings},
  year         = {2025},
  pages        = {115--123}
}

@inproceedings{angeli2015bootstrapped,
  title={Bootstrapped Self Training for Knowledge Base Population.},
  author={Angeli, Gabor and Zhong, Victor and Chen, Danqi and Chaganty, Arun Tejasvi and Bolton, Jason and Premkumar, Melvin Jose Johnson and Pasupat, Panupong and Gupta, Sonal and Manning, Christopher D},
  booktitle={TAC},
  year={2015}
}

@article{sarkhel2023self,
  title={Self-training for label-efficient information extraction from semi-structured web-pages},
  author={Sarkhel, Ritesh and Huang, Binxuan and Lockard, Colin and Shiralkar, Prashant},
  journal={Proceedings of the VLDB Endowment},
  volume={16},
  number={11},
  pages={3098--3110},
  year={2023},
  publisher={VLDB Endowment}
}

@inproceedings{kharrat2023cl_survey,
  author    = {Kharrat, Asma and Drira, Fadoua and Lebourgeois, Franck and Kerautret, Bertrand},
  title     = {Advancements and Challenges in Continual Learning for Natural Language Processing: Insights and Future Prospects},
  booktitle = {Proceedings of the 16th International Conference on Agents and Artificial Intelligence (ICAART 2024)},
  year      = {2024},
  publisher = {SCITEPRESS},
  pages     = {1255--1262},
  volume    = {3},
  doi       = {10.5220/0012662300003663}
}

@article{rostami2024continuous,
  author  = {Rostami, Mohammad},
  title   = {Continuous Unsupervised Domain Adaptation Using Stabilized Representations and Experience Replay},
  journal = {Neurocomputing},
  volume  = {597},
  pages   = {128017},
  year    = {2024},
  doi     = {10.1016/j.neucom.2024.128017}
}

@article{salvador2025review,
  title={A review on knowledge and information extraction from PDF documents and storage approaches},
  author={Salvador, Atagong Desconsciences and TONNANG, Henri Edouard ZEFACK and Odindi, John},
  journal={Frontiers in Artificial Intelligence},
  volume={8},
  pages={1466092},
  year={2025},
  publisher={Frontiers}
}

@article{mruthyunjaya2023rethinking,
  title={Rethinking language models as symbolic knowledge graphs},
  author={Mruthyunjaya, Vishwas and Pezeshkpour, Pouya and Hruschka, Estevam and Bhutani, Nikita},
  journal={arXiv preprint arXiv:2308.13676},
  year={2023}
}

@incollection{mitchell1989theo,
  title={Theo: A framework for self-improving systems},
  author={Mitchell, Tom M and Allen, John and Chalasani, Prasad and Cheng, John and Etzioni, Oren and Ringuette, Marc and Schlimmer, Jeffrey C},
  booktitle={Architectures for intelligence},
  pages={323--355},
  year={1989},
  publisher={Psychology Press}
}

@article{mitchell2018never,
  title={Never-ending learning},
  author={Mitchell, Tom and Cohen, William and Hruschka, Estevam and Talukdar, Partha and Yang, Bishan and Betteridge, Justin and Carlson, Andrew and Dalvi, Bhavana and Gardner, Matt and Kisiel, Bryan and others},
  journal={Communications of the ACM},
  volume={61},
  number={5},
  pages={103--115},
  year={2018},
  publisher={ACM New York, NY, USA}
}

\appendix

\section{Prompts}
The system and user prompts used for DySECT information extraction, are provided in prompts \ref{prompt:System} and \ref{prompt:task-description}, respectively. Moreover, we provide the instruction for injected positive and negative feedback in \ref{prompt:Positive}, and \ref{prompt:Negative}.
\begin{prompt}[title={\footnotesize\texttt{DySECT System Prompt}}, label={prompt:System}]

You are an information extraction agent designed to improve across iterations.

Your goal is to progressively increase recall while maintaining strict precision.

At each iteration, you are given: \\ 
- the document text  \\
- previously extracted triples  \\
- optional additional guidance from earlier iterations  \\

You must treat previously extracted triples as already known facts.
You must not repeat previously extracted triples.

You should look for:  \\
- relations that were missed earlier  \\
- entities that were previously unseen  \\
- new relations involving known entities  \\
- implicit but explicitly stated facts that can be expressed independently  \\

You must not hallucinate, infer unstated facts, or relax schema constraints.  
You must strictly obey the allowed concept types, relations, and output format. \\

Your objective is to extract new, valid triples that increase coverage of the document.

\end{prompt}

\begin{prompt}[title={\footnotesize\texttt{DySECT User Prompt}}, label=prompt:task-description]
\textbf{Task Description}

You are performing ONE ITERATION of knowledge extraction.

Your task is to extract new knowledge triples from the document that were not extracted in previous iterations.

Each triple must represent a self-contained, explicitly stated fact from the text and must stand on its own without relying on surrounding context. \\

\textbf{Extraction Rules}

- A triple represents a factual relationship expressed in the text.  
- The subject is the entity performing or undergoing the action.  
- The object is the factual target of the relationship.  
- Each triple must contain exactly one subject entity and exactly one object entity.  
- Ignore vague, speculative, or non-factual statements.  
- Ignore relations not explicitly stated in the text.  
- Ignore relations not in the allowed allowlist. \\

\textbf{Iteration Constraints (Critical)}

- You are given Previously Extracted Triples (except in the first iteration).  
- Do not output any triple that is semantically identical to an existing one.  
- You may extract new relations between known entities.  
- You may extract relations involving newly discovered entities.  
- Prefer triples that increase document coverage rather than paraphrasing known facts. \\

\textbf{Output Format (Must Follow Exactly)}

- Output in JSON format.  
- Provide a list of items.  
- Each item must follow: (subject, subject\_type, relation, object, object\_type).  
- Output only the extractions.  
- Do not include headers, comments, explanations, or formatting characters. 

Example:
{\ttfamily\small
[
\newline
\quad ["Mount Bailey", "LOC", "country", "U.S.", "LOC"],
\newline
\quad ["Thomas Wolff", "PER", "date of birth", "July 14, 1954", "TIME"]
\newline
]
}
\newline

\textbf{Allowed Concepts and Relations (Strict)}

You must only extract triples where:  \\
1) SubjectType and ObjectType in AllowedConceptTypes  \\
2) relation in AllowedRelations  \\

\textbf{AllowedConceptTypes:} \\
\{docred\_concept\_type\_list\}

\textbf{AllowedRelations:} \\
\{docred\_relations\_list\} \\

\{kb-info\} \\
\newline
\textbf{Document Text}

\begin{verbatim}
{document}
\end{verbatim}
\end{prompt}

\begin{prompt}[title={\footnotesize\texttt{Encouraging (Positive)}}, label=prompt:Positive]
\textbf{Previously Extracted Triples}

You are given extractions from the previous iteration.  
Treat these as positive examples.

Use them to guide your reasoning and extract additional entities and relations of similar types and patterns from the document.

Do not repeat the same triples unless strictly necessary.  
Focus on finding new, previously unextracted information.

\begin{verbatim}
{example}
\end{verbatim}

\{added\_info\}
\end{prompt}

\begin{prompt}[title={\footnotesize\texttt{Prohibitive (Negative)}}, label=prompt:Negative]
\textbf{Previously Extracted Triples}

You are also given a set of examples from previous iterations.
These represent entity and relation patterns that are either irrelevant, low-value, or already over-extracted.

Use them as signals for what NOT to focus on.
Avoid extracting similar entities or relations.
Instead, prioritize discovering new types of information that are not covered by these examples.

\begin{verbatim}
{example}
\end{verbatim}

\{added\_info\}
\end{prompt}

\section{System Demonstration Extra Features}
\begin{figure*}[t!]
  \centering
  \includegraphics[width=.8\textwidth]{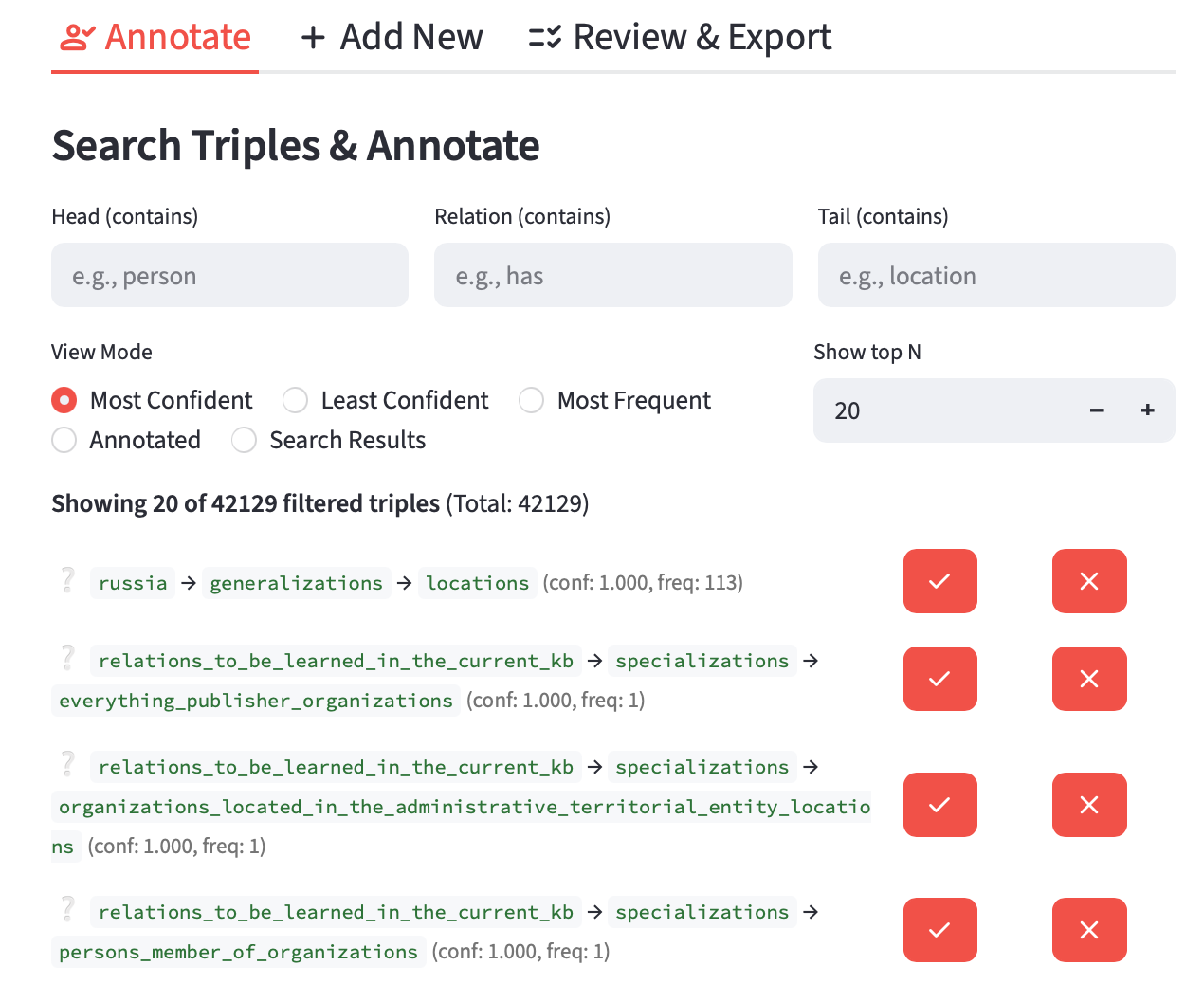}
  \caption{Annotation dashboard allowing users to inspect, validate, or reject extracted triples within the knowledge base.}
  \label{fig:annotate}
\end{figure*}

\begin{figure*}[t]
  \centering
  \begin{subfigure}{0.45\textwidth}
    \includegraphics[width=\textwidth]{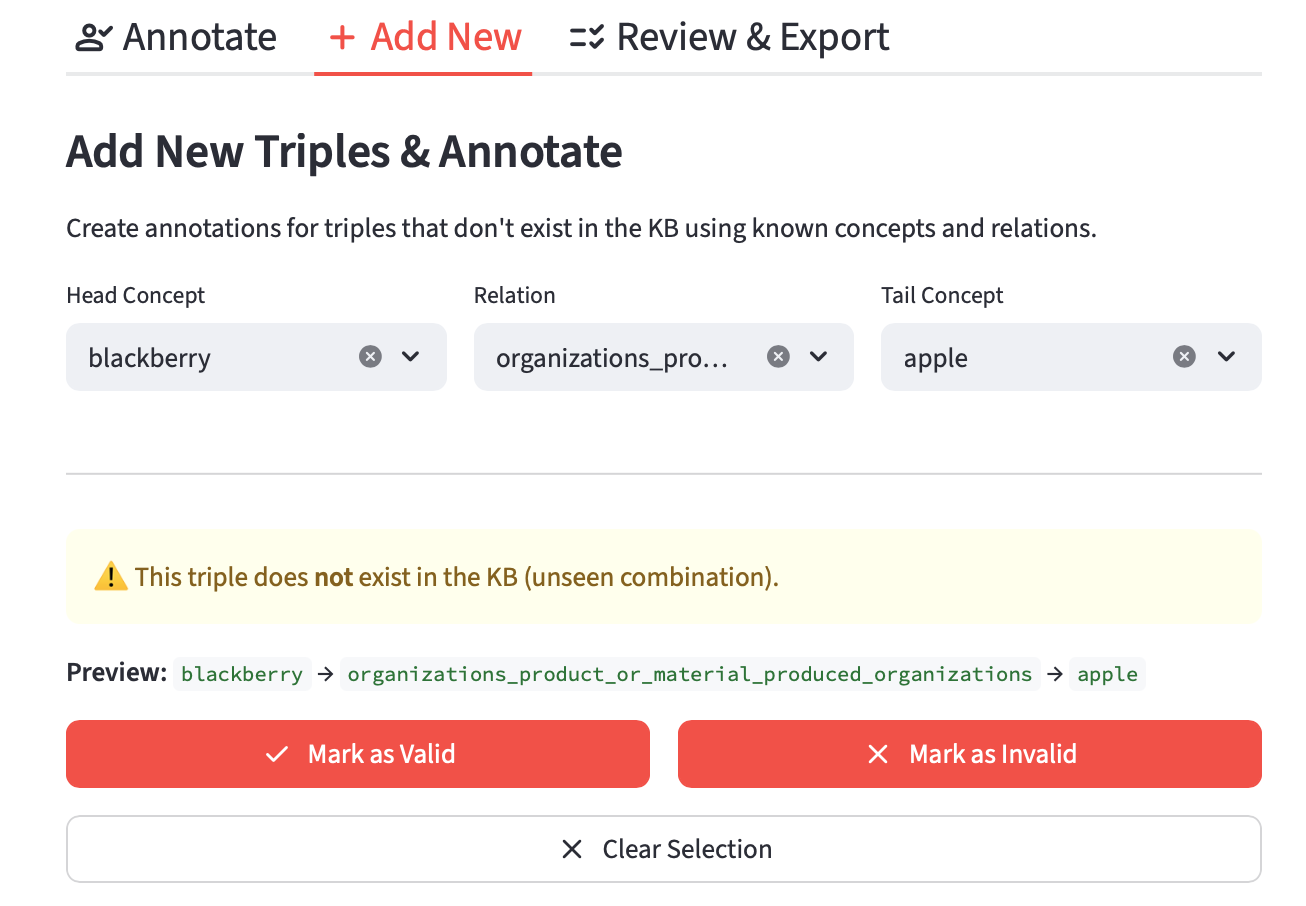}
    \caption{Interface for manually adding or correcting triples to enforce policies, mitigate bias, or refine the knowledge base.}
    \label{fig:add}
  \end{subfigure}
  \hfill 
  \begin{subfigure}{0.45\textwidth}
    \includegraphics[width=\textwidth]{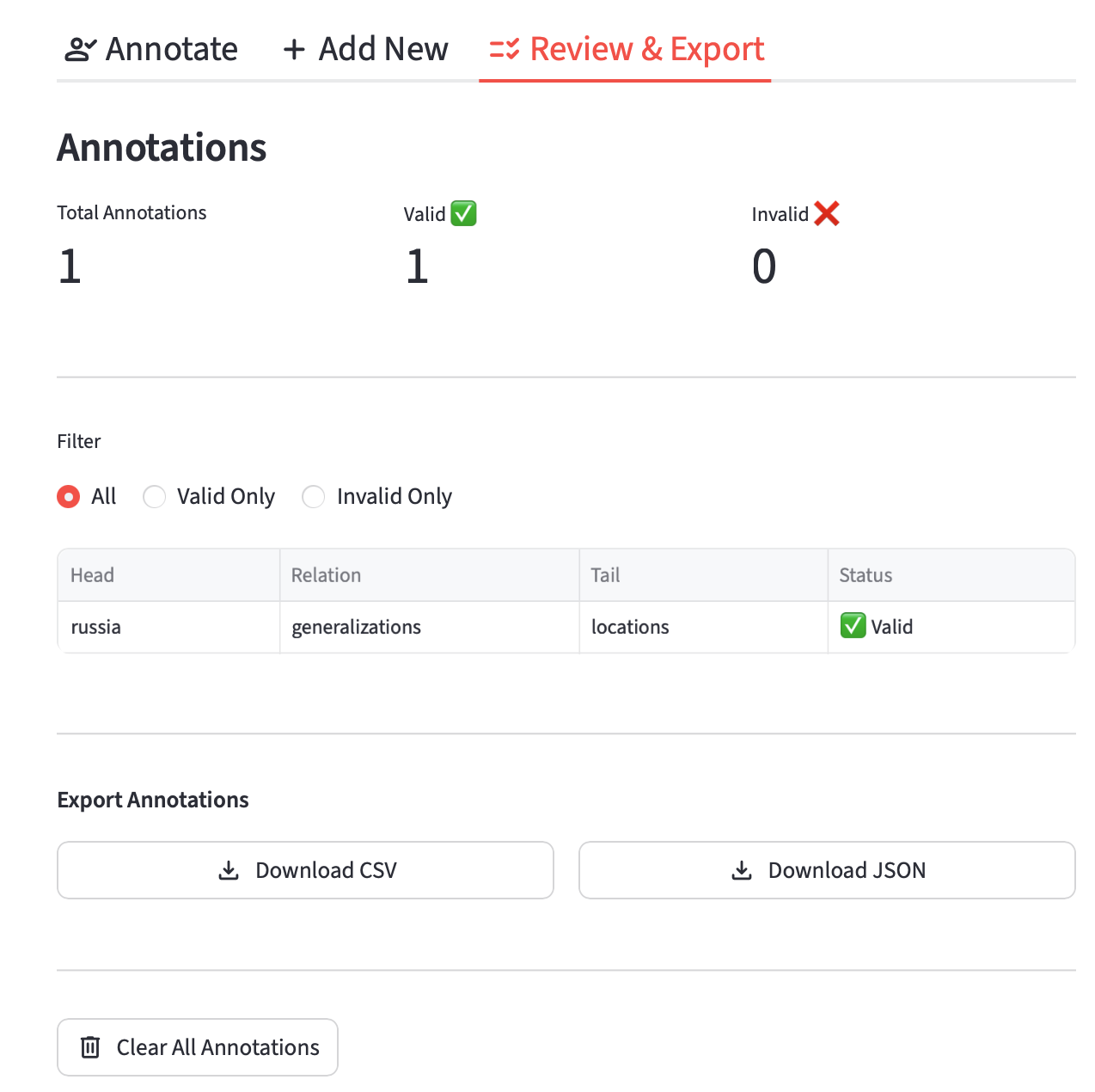}
    \caption{Annotation review interface for auditing and exporting validated triples.}
    \label{fig:annotate-review}
  \end{subfigure}
  \caption{Triple addition, review, and reporting sections.}
  \label{fig:main}
\end{figure*}

We provide an annotation dashboard for manual validation and invalidation of extracted triples, shown in Figure~\ref{fig:annotate}. The ``Add New'' section (Figure~\ref{fig:add}) allows users to introduce new relations between concepts or modify their validation status, while the ``Review \& Export'' section provides an overview of annotation activities and high-level KB statistics (Figure~\ref{fig:annotate-review}).

\begin{figure*}[th!]
  \centering
  \includegraphics[width=\textwidth]{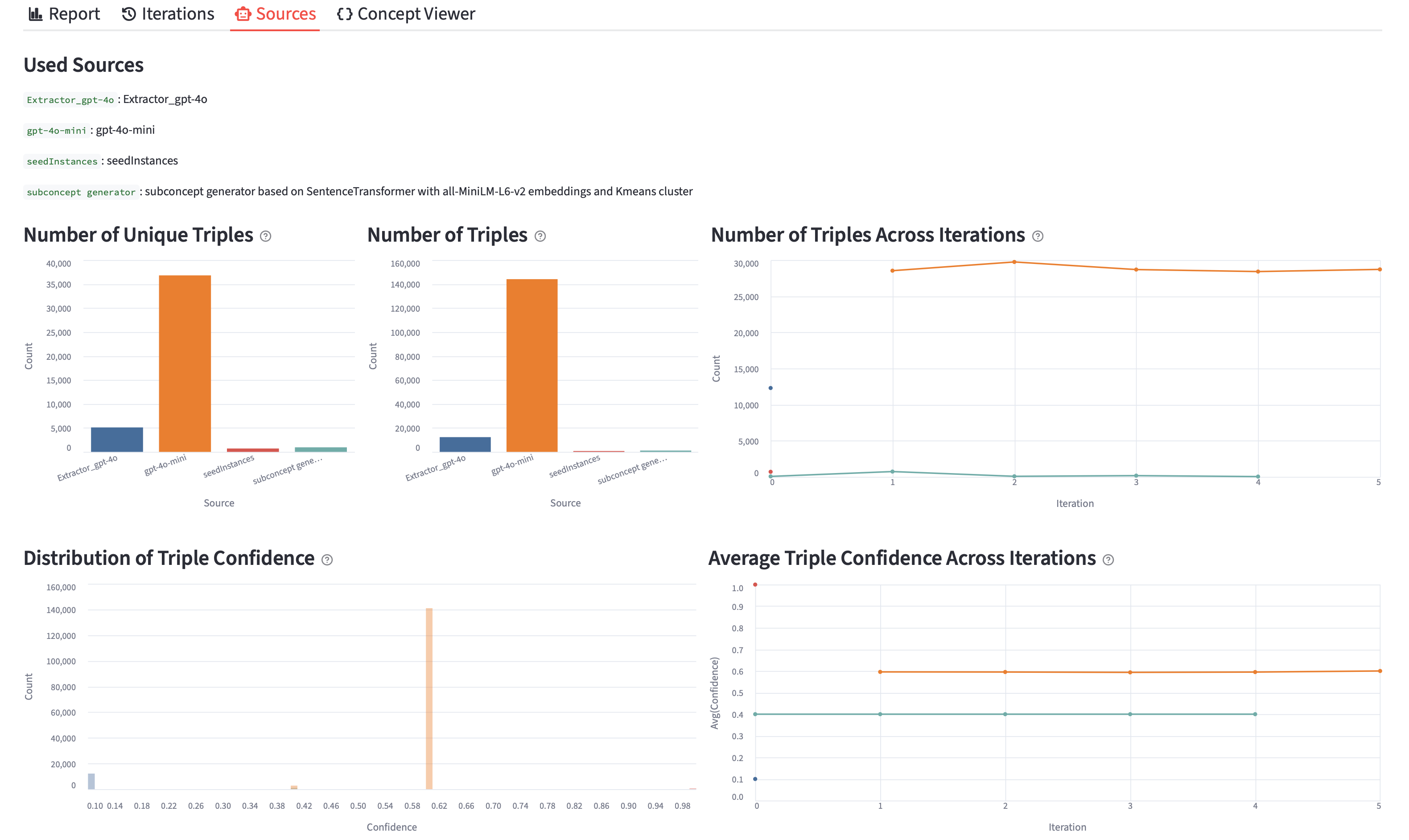}
  \caption{Overview of triple provenance, showing the distribution of extracted triples by source.}
  \label{fig:sources}
\end{figure*}

We can monitor source statistics and confidence trends across extraction iterations through the ``Sources'' section shown in Figure~\ref{fig:sources}.

\begin{figure*}[t]
  \centering
  \includegraphics[width=0.75\textwidth]{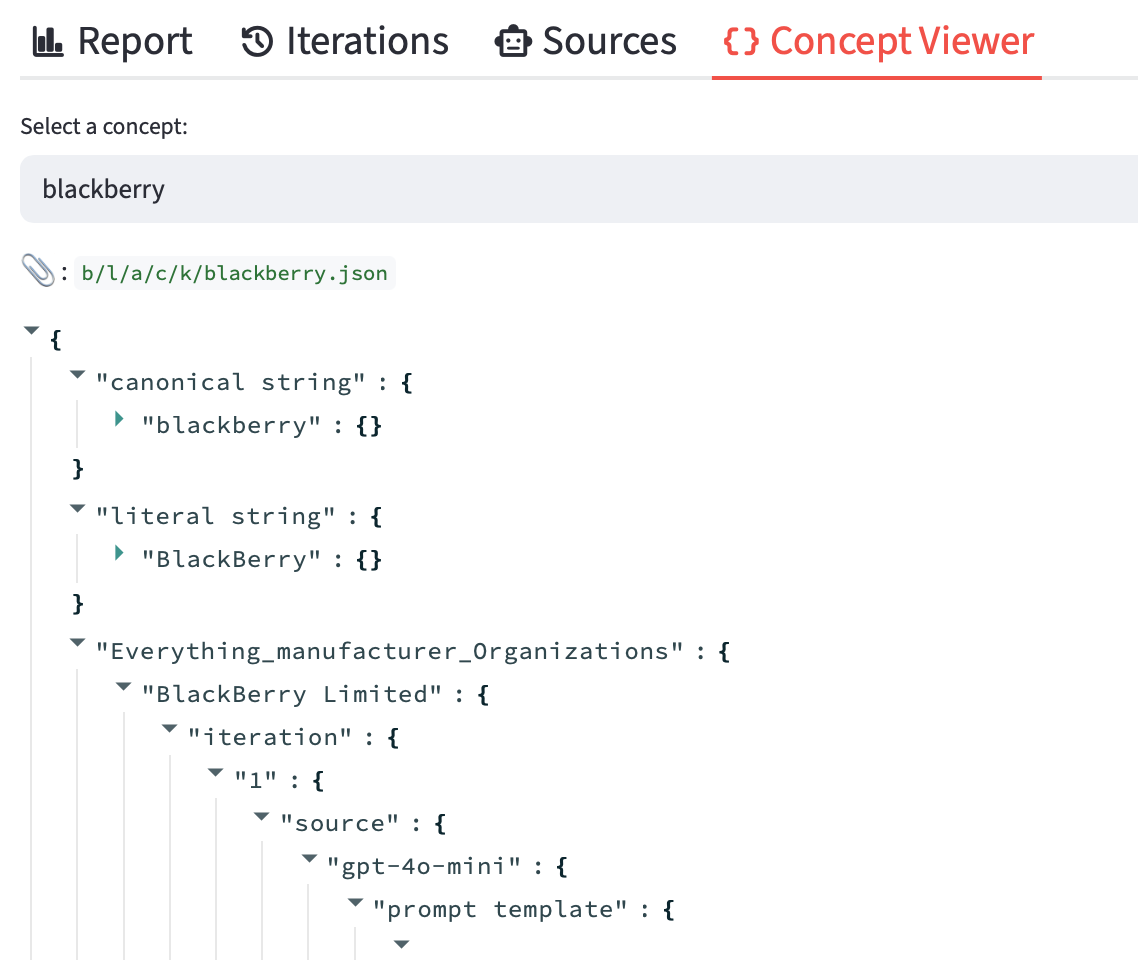}
  \caption{Concept Viewer interface displaying the raw stored representation of a selected concept in the knowledge base.}
  \label{fig:viewer}
\end{figure*}

Finally, the ``Concept Viewer'' provides direct access to the raw KB contents associated with individual concepts and relations, as shown in Figure~\ref{fig:viewer}.

\begin{figure*}[t]
  \centering
  \includegraphics[width=\textwidth]{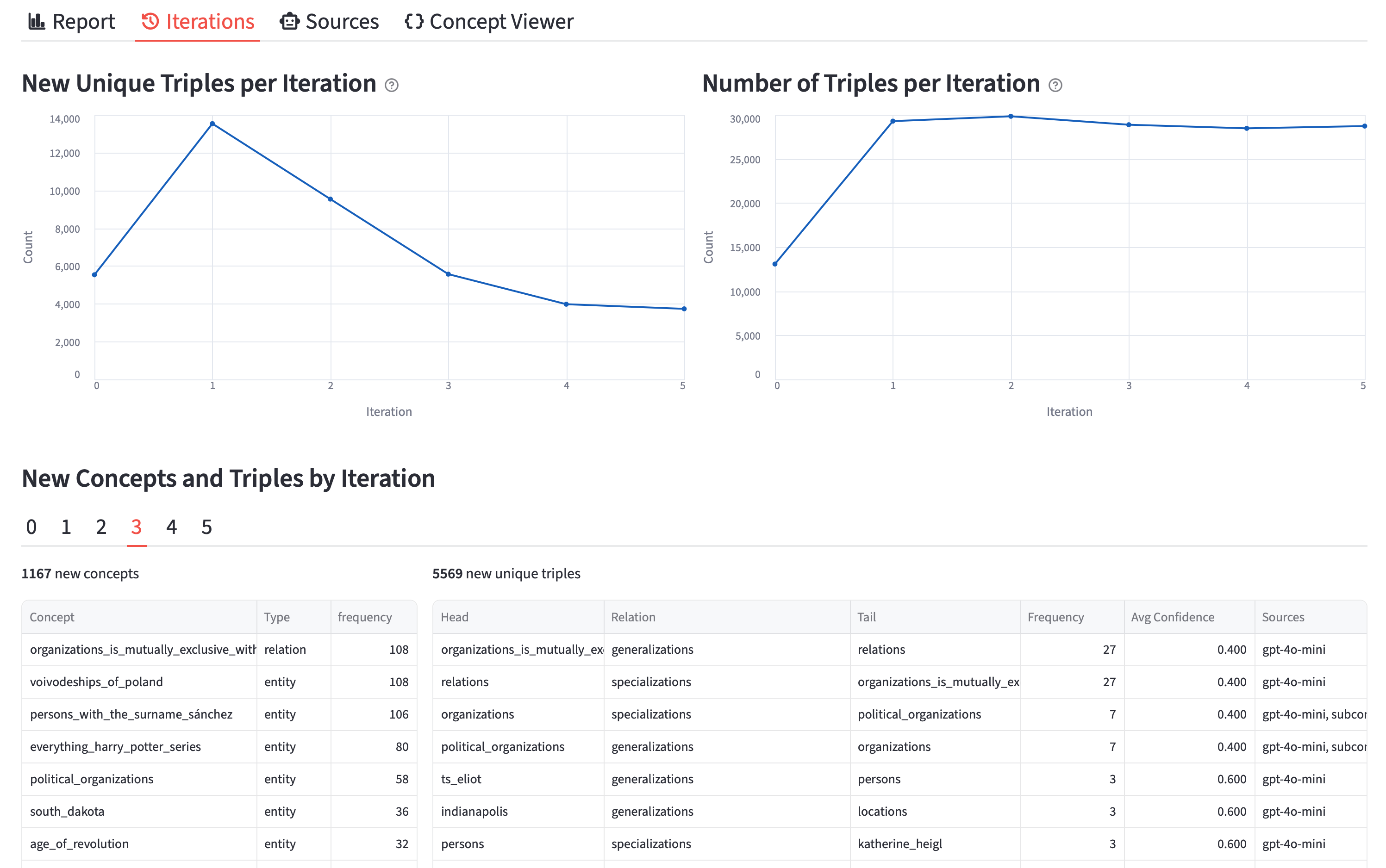}
  \caption{Evolution of triples and concepts across iterations, including frequency and confidence summaries.
}
  \label{fig:iterations}
  \minipostspace
\end{figure*}

\begin{figure*}[t]
  \centering
  \includegraphics[width=0.75\linewidth]{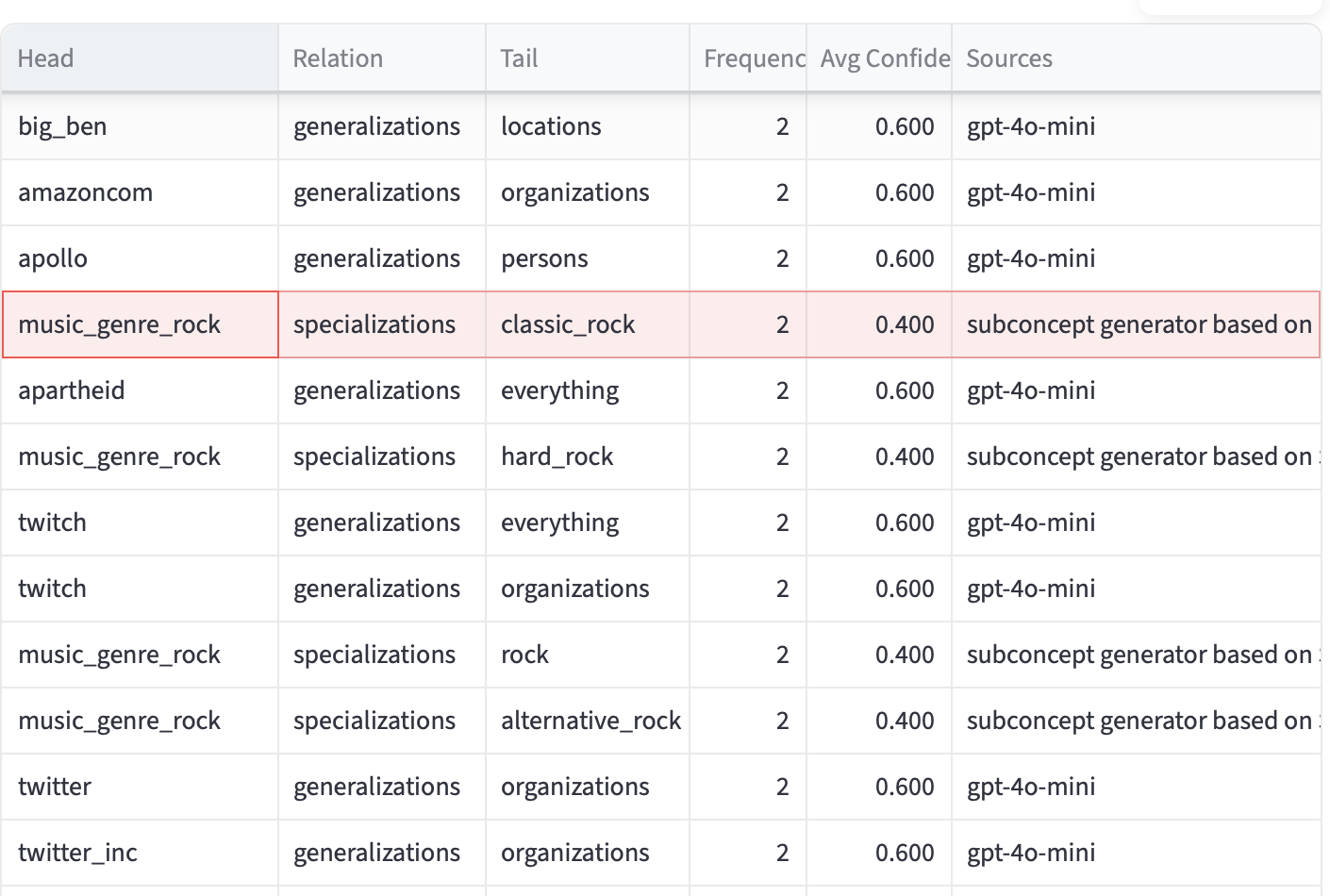}
  \caption{Emergence of the new subconcept (\textit{Music Genre: Rock}) in the first knowledge integration iteration with associated source and confidence.}
  \label{fig:emerg}
  \minipostspace
\end{figure*}

\begin{figure*}[t]
  \centering
  \includegraphics[width=\textwidth]{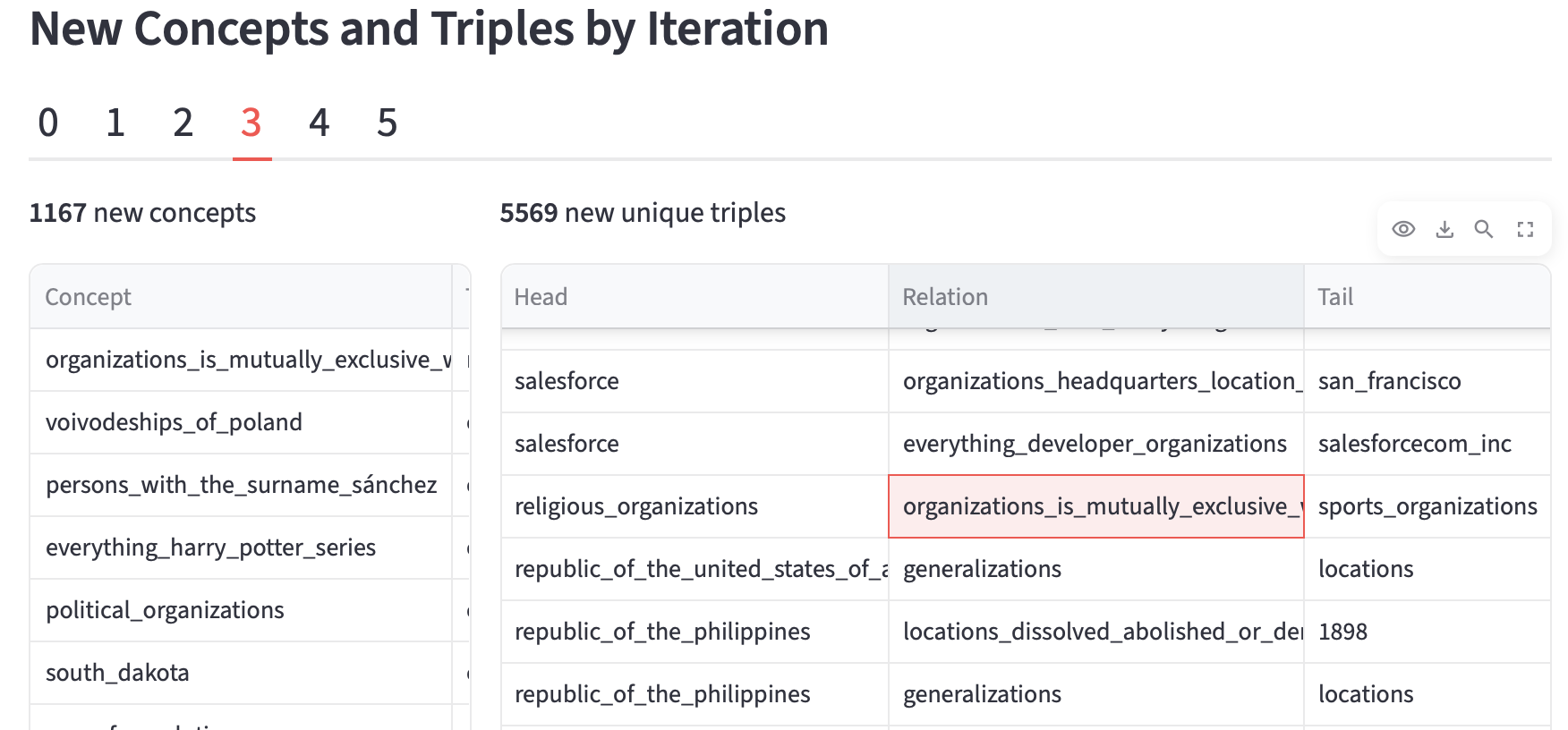}
  \caption{Identification of mutually exclusive concept groups during the third knowledge integration inner-loop iteration.
}
  \label{fig:mutex}
  \postspace
\end{figure*}

\end{document}